\documentclass[11pt,a4paper]{article}
\usepackage[hyperref]{emnlp2018}
\usepackage{times}
\usepackage{latexsym}
\usepackage{amsmath}
\usepackage{url}
\usepackage{graphicx}
\usepackage{float}
\aclfinalcopy 

\title{Attentive Semantic Role Labeling with Boundary Indicator}

\author{Zhuosheng Zhang$^{1,2}$, Shexia He$^{1,2}$, Zuchao Li$^{1,2}$, Hai Zhao$^{1,2,}$\thanks{$\ $ Corresponding author.}\\
	$^{1}$Department of Computer Science and Engineering, Shanghai Jiao Tong University \\
	$^{2}$Key Laboratory of Shanghai Education Commission for Intelligent Interaction \\ and Cognitive Engineering, Shanghai Jiao Tong University, Shanghai, 200240, China\\
	{\tt \{zhangzs, heshexia, charlee\}@sjtu.edu.cn, {\tt zhaohai@cs.sjtu.edu.cn}}
}

\date{}

\begin{document}
\maketitle
\begin{abstract}

\end{abstract}
The goal of semantic role labeling (SRL) is to discover the predicate-argument structure of a sentence, which plays a critical role in deep processing of natural language. This paper introduces simple yet effective auxiliary tags for dependency-based SRL to enhance a syntax-agnostic model with multi-hop self-attention. Our syntax-agnostic model achieves competitive performance with state-of-the-art models on the CoNLL-2009 benchmarks both for English and Chinese. 

\section{Introduction}

Semantic role labeling (SRL) aims to derive the meaning representation for a sentence, i.e., predicate-argument structure, which plays a critical role in a wide range of natural language processing tasks \cite{2018_aaai_semanticilp,Huang2018Moon,zhu2018lingke,zhang2018OneShot,zhang2018char}. There are two formulizations for semantic predicate-argument structure, one is based on constituents (i.e., phrase or span), the other is based on dependencies. The latter is also called semantic dependency parsing, which annotates the heads of arguments rather than phrasal arguments. SRL can be formed as four subtasks, including \emph{predicate detection}, \emph{predicate disambiguation}, \emph{argument identification} and \emph{argument classification}. 

Recent methods \cite{zhou-xu2015,marcheggiani2017,marcheggianiEMNLP2017,he-acl2017,selfatt2018,He2018Syntax,C18-1271,He2018Unified} deal with all words in entire sentence instead of distinguishing arguments and non-arguments which actually differ in quantity. The indiscriminate treatment would result in a serious unbalanced issue for argument labeling. 

We observe that arguments trend to surround their predicates. Capturing the boundary of the semantic relationship beforehand and taking it as an inference constraint is thus particularly significant for argument labeling, which is potential to improve the performance of the labeler. In this work, we propose to introduce two types of auxiliary argument tags as the argument boundary indicators. If an argument candidate is assigned to such either of the tags, the labeling or traversal algorithm will end immediately. In training, this auxiliary tags mean no more samples will be searched for the current predicate, while in inference, the labeler will not search arguments any more. The auxiliary tags could guide the labeler to focus on the potential true candidates. 

Besides, most of state-of-the-art models rely heavily on syntactic features \cite{roth2016,marcheggianiEMNLP2017} which suffer the risk of erroneous syntactic input, leading to undesired error propagation. In
fact, there comes a latest advance that shows neural SRL able to effectively capture the discriminative information automatically without syntactic assistance \cite{marcheggiani2017}. Furthermore, for long and complex sentences with various aspects of semantics, effectively modeling the overall sentence would be quite challenging. To this end, we introduce a multi-hop self-attention mechanism to distill various important parts of the input sentence and model long range dependencies.

This paper focuses on argument identification and classification, which is jointly formulized  as a sequence labeling task. For the predicate disambiguation, we follow the previous works \cite{roth2016,marcheggiani2017}. Our model contains two major features: (1) auxiliary tags to indicate the argument boundary. (2) a BiLSTM encoder with multi-hop self-attention to model the sentence representations. Our evaluation is on CoNLL-2009 \cite{hajivc-EtAl2009} benchmark for both English and Chinese. we show that with the help of auxiliary tags and self-attention, the syntax-agnostic model could even achieve a competitive performance with syntax-aware ones. 

\section{Argument Boundary Indicator}

\begin{table}\small
	\centering
	\setlength{\tabcolsep}{3pt}
	\begin{tabular}{lccc|ccc}
		\hline	
		
		\hline
		& \multicolumn{3}{c|}{Without AT (\%)} & 	\multicolumn{3}{c}{With AT (\%)} \\
		 & Args & NonArgs & Ratio & Args & NonArgs & Ratio\\
		 	\hline
		 Train & 7.65 & 92.35 & 1:13 & 43.81 & 56.19 & 1:1.3\\
Dev & 7.35  & 92.65 & 1:13 & 41.22 & 58.78 & 1:1.4\\
		\hline	
		
		\hline
	\end{tabular}
	\caption{Label distribution of training and dev set. \emph{Arg} is short for argument. \emph{AT} denotes our introduced auxiliary tags. }\label{tab:data_dist}
\end{table}

Following the observation that arguments usually tend to surround their predicate closely, we introduce two auxiliary tags inspired by \cite{zhao2009}, namely, \emph{beginning of the argument, $<$BOA$>$} and \emph{end of the argument, $<$EOA$>$} to signify where the labeler should start or stop collecting argument candidates. For training, both tags are correspondingly assigned to the previous or next word as soon as the arguments of the current predicate have been saturated with previously collected words, in light of the original training data. For inference, it informs the labeler to start argument searching when it comes to the $<$BOA$>$ while $<$EOA$>$ means to stop. These tags would help the  labeler ignore those words too far away from the predicates which are hardly supposed to be ground-truth arguments. 

Empirically, the distributions of arguments (Args) and non-arguments (NonArgs) vary largely in quantity. Table \ref{tab:data_dist} shows the data statistics of CoNLL 2009 dataset for English and we find the proportion of Args and NonArgs is 1:13 in the original dataset. After replacing the semantic relationship boundary (both left and right) with our new tags and removing all other NonArg labels, the proportion reaches nearly 1:1. Note that the above operation is only conducted to intuitively show the difference by imitating the enhanced searching guidance with new tags. Actually we only modify the boundary labels of semantic relationships and use them to signal the model where to restrict a search. Without this inference restraint, most argument candidates are irrelevant and far away from the current predicate, inevitably interfering with the informative features from the truly relevant ones in the very small minority and, hence, leading to an unsatisfactory performance.

We give an example below to show how these two tags are used. Suppose a sequence with  sense-disambiguated predicate and labeled arguments is
\begin{table}[H]\footnotesize
	\centering
	\begin{tabular}{ c c c c c c c}
		a & big &  apple &  \textbf{drops} & from & the & tree \\
		\_ & \_ &  A1 &  \_ & A3 & \_ & \_ \\
	\end{tabular}
\end{table}   
\noindent where \emph{drops} in the input sequence is a predicate with two arguments, labeled with $A0$ and $A1$, respectively. 

The two tags are assigned to the next two words \emph{apple} and \emph{from}, respectively, indicating no more arguments farther than them from the predicate. 

\begin{table}[H]\footnotesize
	\centering
	\begin{tabular}{ c c | c c c | c c}
		a & big &  apple &  \textbf{drops} & from & the & tree \\
		\_ & $<$BOA$>$ &  A1 &  \_ & A3 & $<$EOA$>$ & \_ \\
	\end{tabular}
\end{table} 

The auxiliary tags can be regarded as a reference constraint which indicates the maximum boundary of the argument set for each predicate. They are treated in exactly the same way as all other labels during training and inference, except the extra utility to signal where to stop a search during decoding inference.
 \begin{figure}
 	\centering
 	\includegraphics[width=0.5\textwidth]{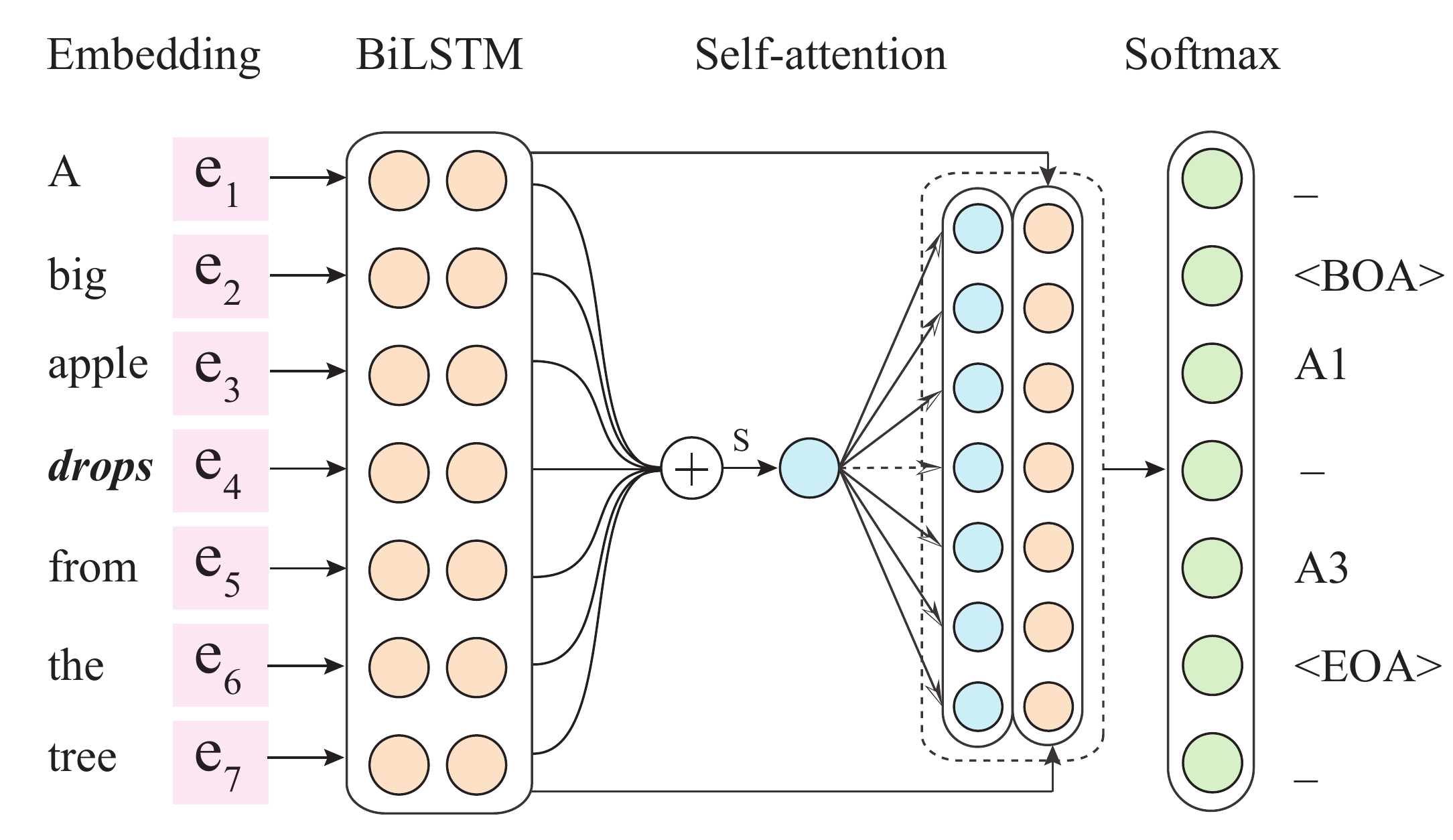}
 	\caption{\label{fig:overview} Model architecture.}
 \end{figure}
\section{Bidirectional LSTM Labeler}
Figure \ref{fig:overview} overviews our model architecture. Given a known predicate, our model reads each word of an input sentence and maps it into latent space to form a word-level representation. Concretely, each word embedding is defined by
\begin{align*}
e_i = [e_i^{r}, e_i^{p}, e_i^{l}, e_i^{pos}, e_i^{f}, e_i^{m}]
\end{align*}
where $e_i^{r}$ is randomly initialized word embedding, $e_i^{p}$ denotes pre-trained word embedding, $e_i^{l}$ represents randomly initialized lemma embedding, $e_i^{pos}$ is the randomly initialized POS tag embedding and $e_i^{f}$ denotes predicate-specific indicator embedding to indicate whether the current word is the given predicate, which is slightly different from previous work \cite{marcheggiani2017} directly using a binary flag either 0 or 1 and $e_i^{m}$ is an external embedding, ELMo (Embeddings from Language Models) \cite{ELMo}, which is obtained by deep bidirectional language model that takes characters as input. 

The concatenated word embeddings $e_i$ are then fed to a sentence-level module to propagate information along the input sequence. We use a bidirectional LSTM (BiLSTM) \cite{Hochreiter1997} to process the sequence $e = (e_1,\dots,e_n)$ in forward and backward directions to access both past and future contextual information. Finally, we get a contextual representation $h_i = [\overrightarrow{h}_i, \overleftarrow{h}_i]  \in R^{n \times 2d}$ where $d$ denotes the number of LSTM hidden units. $\overrightarrow{h}_i$ denotes the hidden states of the sequence from $e_1$ to $e_i$ and $\overleftarrow{h}_i$  represent that from $e_n$ to $e_i$ . 

Attention mechanism has been applied to a wide range of tasks due to its effectiveness of key information extraction \cite{Lin2017A,zhang2018SubMRC,zhang2018DUA}. To pinpoint important components of the sentence, such as critical words or phrases, we employ a self-attention mechanism  following \cite{Lin2017A} to obtain a vector of weights $m$.
\begin{align*}
m = softmax(W_2\tan(W_1H^T))
\end{align*}
where $W_1 \in R^{k \times 2d}$ and $W_2  \in R^{k}$ are model parameters where $k$ is an arbitrary hyper-parameter. In this work, we empirically set $k=d$. Then we sum up the BiLSTM hidden states $H=(h_1,\dots,h_n)$ weighted by $m$ to obtain an attentive representation $s$ of the whole input sentence.

In fact, there might be multiple aspects or semantic components of a sentence, especially for a long sentence. Thus, we need multiple $m$ to focus on different parts of the sentence, which lets us adopt multi-hop attention. Let $r$ denote the number of different parts to be extracted from the sentence, we expand $W_2$ into $r$ dimension, thus we have $W_2 \in R^{r \times 2d}$ and the resulting weight vector $m$ becomes a matrix $M$. Then, we compute the weighted sums by multiplying $M$ and BiLSTM hidden states $H$ to obtain the multi-hop attentive sentence representation $S=MH$. Intuitively, multi-hop self-attention provides a flexible way to represent, extract and synthesize diverse information of input sequence which would produce a more fine-grained global sentence information.

Then we concatenate each hidden state $h_i$ with $S$ to endow each word representation with contextual sentence information. Here, we have the refined output $\hat{H}=\left [ h_1 \diamond S; h_2 \diamond S; \cdots ;h_n \diamond S; \right ]$ where $\diamond$ denotes concatenation operation.

Eventually, we use a softmax layer over $\hat{H}$. The training objective is to maximize the logarithm of the likelihood of the labels.
\begin{align*}
\ell = - \sum_{i=1}^{n} y_{i}\log \hat y_i
\end{align*}
where $y_{i}$ denotes the prediction, $\hat{y}_{i}$ is the target. During inference, we use greedy search to obtain the prediction. Note the search start from the predicate with two directions, forward and backward, until the argument boundary tag is predicted.
\section{Experiment}

\begin{table}
	\centering
	\setlength{\tabcolsep}{3pt}
	\begin{tabular}{lccc}
		\hline
		
		\hline
		System (syntax-aware) & P & R & F$_1$ \\
		\hline
		\textit{Single model} & & &\\
		\citet{bjorkelund2010} & 87.1 & 84.5 & 85.8 \\
		\citet{Lei2015} & $-$ & $-$ & 86.6 \\
		\citet{Fitzgerald2015} & $-$ & $-$ & 86.7 \\
		\citet{roth2016} & 88.1 & 85.3 & 86.7\\
		\citet{marcheggianiEMNLP2017} & 89.1 & 86.8 & 88.0 \\
		\hline
		\textit{Ensemble model} & & &\\
		\citet{Fitzgerald2015} & $-$ & $-$ & 87.7 \\
		\citet{roth2016} & 90.3 & 85.7 & 87.9\\
		\citet{marcheggianiEMNLP2017} & 90.5 & 87.7 & 89.1 \\
		\hline
		System (syntax-agnostic) & P & R & F$_1$ \\
		\hline
		\citet{marcheggiani2017} & 88.7 & 86.8 & 87.7 \\
		\textbf{Ours} & \textbf{89.7} & \textbf{88.3} & \textbf{89.0} \\
		\hline
		
		\hline
	\end{tabular}
	\caption{Results on the English in-domain test set.}\label{tab:resultE}
\end{table}

Our model is evaluated on the CoNLL-2009 shared task both for English and Chinese datasets, following the standard training, development and test splits. In our experiments, the pre-trained word embeddings for English are 100-dimensional GloVe vectors \cite{penningtonEMNLP2014}. For Chinese, we exploit Wikipedia documents to train the same dimensional Word2Vec embeddings \cite{Mikolov2013}. All other vectors are randomly initialized, the dimensions of word and lemma embeddings are 100, while the dimensions of POS tag and predicate indicator embedding are 32 and 16 respectively. In addition, we use 300-dimensional ELMo embedding for English. For multi-hop self-attention, we set $r = 10$. Our evaluation is based on the following metrics: Precision (P), Recall (R) and F$_1$-score.

During training procedures, we use the categorical cross-entropy as objective, with Adam optimizer \cite{adam2015} with learning rate 0.001, and the batch size is set to 64. The BiLSTM encoder consists of 4 BiLSTM layers with 512-dimensional hidden units. We apply dropout for BILSTM with a 90\% keep probability between time-steps and layers. We train models for a maximum of 20 epochs and obtain the nearly best model based on development results.

\subsection{Results}
The experimental results on the in-domain English data and Chinese test set are in Tables \ref{tab:resultE} and \ref{tab:resultC}, respectively.
Notably, our syntax-agnostic model is local (argument identification and classification decisions are conditionally independent) and single without reranking, which neither includes global inference nor combines multiple models.
\begin{table}
	\centering
	\setlength{\tabcolsep}{3pt}
	\begin{tabular}{lccc}
		\hline
		
		\hline
		System (syntax-aware) & P & R & F$_1$ \\
		\hline
		\citet{bjorkelund2009} & 82.4 & 75.1 & 78.6 \\
		\citet{roth2016} & 83.2 & 75.9 & 79.4 \\
		\citet{marcheggianiEMNLP2017} & 84.6 & 80.4 & 82.5 \\
		\hline
		System (syntax-agnostic) & P & R & F$_1$ \\
		\hline
		\citet{marcheggiani2017} & 83.4 & 79.1 & 81.2 \\
		\textbf{Ours} & \textbf{84.3} & \textbf{79.6} & \textbf{81.9} \\
		\hline
		
		\hline
	\end{tabular}
	\caption{Results on the Chinese test set.}\label{tab:resultC}
\end{table}

\begin{table}
	\centering
	\setlength{\tabcolsep}{3pt}
	\begin{tabular}{lccc}
		\hline
		
		\hline
		System (syntax-aware) & P & R & F$_1$ \\
		\hline
		\textit{Single model}&&&\\
		\citet{bjorkelund2010} & 75.7 & 72.2 & 73.9 \\
		\citet{Lei2015} & $-$ & $-$ & 75.6 \\
		\citet{Fitzgerald2015} & $-$ & $-$ & 75.2 \\
		\citet{roth2016} & 76.9 & 73.8 & 75.3 \\
		\citet{marcheggianiEMNLP2017} & 78.5 & 75.9 & 77.2 \\
		\hline
		\textit{Ensemble model}&&&\\
		\citet{Fitzgerald2015} & $-$ & $-$ & 75.5 \\
		\citet{roth2016} & 79.7 & 73.6 & 76.5 \\
		\citet{marcheggianiEMNLP2017} & 80.8 & 77.1 & 78.9 \\
		\hline
		System (syntax-agnostic) & P & R & F$_1$ \\
		\hline
		\citet{marcheggiani2017} & 79.4 & 76.2 & 77.7 \\
		\textbf{Ours} & \textbf{81.5} & \textbf{76.1} & \textbf{78.7} \\
		\hline
		
		\hline
	\end{tabular}
	\caption{Results on the English out-of-domain test set.}\label{tab:resultood}
\end{table}

For English, as shown in Table \ref{tab:resultE}, our  model outperforms previously published single models including syntax-aware models, scoring 89.0\% F$_1$ with 1.3\% absolute improvement over the syntax-agnostic baseline in the in-domain test set. 

For Chinese (Table \ref{tab:resultC}), even though we use the same hyper-parameters as for English, our model also shows competitive performance with state-of-the-art results. Table \ref{tab:resultood} demonstrates the results on out-of-domain data, where the performance of our model still remains strong.

\section{Analysis}
Result \ref{tab:ana} shows the ablation study of our model which indicates all our proposed strategies contribute to the overall performance. Without auxiliary tags, the performance drops dramatically, which confirms the soundness of the motivation for argument boundary indicators from empirical perspective. The reason might be that our proposed argument boundary indicators could help the labeler focus on the potential true candidates and ignore those words too far away from the predicates which are hardly supposed to be ground-truth arguments. Removing the self-attention module also results in performance decline, the advance might be because the self-attention mechanism could help the model to distill vital information and alleviate the error propagation.

\begin{table}
	\centering
	\begin{tabular}{lccc}
		\hline	
		
		\hline
		System & P & R & F$_1$ \\
		\hline
		Ours & \textbf{89.7} & \textbf{88.3} & \textbf{89.0}  \\
		-Auxiliary tags	& 89.5  & 88.1  & 88.8 \\
		-Self-attention	& 89.7  & 87.9  &  88.7 \\
		-Auxiliary tags -self-attention	& 88.9  & 88.1  & 88.5  \\
		\hline
		+Adaptive argument pruning & 88.6 & 85.5 & 87.0\\
		\hline
		
		\hline
	\end{tabular}
	\caption{\label{tab:ana}Results on the English in-domain test set.}
\end{table}

Noting that the work  \cite{zhao-jair-2013} successfully incorporated the syntactic information by applying an adaptive argument pruning, we further perform an experiment to explore whether employing such pruning method enhance or hinder our model.
However, as shown in Table \ref{tab:ana}, the
result is far from satisfying.

\section{Conclusion}
This paper introduced auxiliary tags to indicate the boundary of predicate-argument relationships and employ multi-hop self-attention for further improvement of SRL performance. With the auxiliary tags and the attention mechanism, our simple yet effective model achieves competitive results on the CoNLL-2009 benchmarks for both English and Chinese, though without any kind of syntactic information.

\bibliography{emnlp2018}
\bibliographystyle{acl_natbib_nourl}

\end{document}